%% file: root.tex
\documentclass[letterpaper,10pt,conference]{ieeeconf}

\IEEEoverridecommandlockouts
\usepackage{hyperref}
\usepackage{amssymb}
\usepackage{amsmath}
\usepackage{amsfonts}
\usepackage{mathptmx}
\usepackage{times}
\usepackage{booktabs}
\usepackage{multirow}
\usepackage{placeins}
\usepackage{afterpage}
\usepackage{graphicx}
\usepackage{caption}
\usepackage{xspace}
\usepackage{color,soul}

\usepackage[backend=biber,style=ieee,maxbibnames=6,minbibnames=1,url=false,doi=false,isbn=false]{biblatex}
\AtEveryBibitem{%
  \clearfield{pages}\clearfield{volume}\clearfield{number}\clearfield{month}%
  \clearfield{series}\clearfield{edition}\clearfield{note}\clearfield{issn}%
  \clearfield{eprintclass}\clearlist{publisher}\clearlist{location}\clearlist{address}%
  \clearname{editor}\clearfield{urldate}%
}
\hypersetup{
  pdftitle={When Does Touch Matter? Charting the Vision-Interaction Gap in Cluttered Dexterous Grasping},
  pdfauthor={Hao Jiang, Luis Dominguez, Daniel Seita},
  pdfsubject={Robotics},
  pdfkeywords={dexterous manipulation, tactile sensing, force sensing, cluttered grasping, imitation learning}
}

\newcommand{\ba}{\mathbf{a}}

\newcommand{\bo}{\mathbf{o}}
\newcommand{\bs}{\mathbf{s}}

\definecolor{teal}{rgb}{0.0,0.5,0.5}
\definecolor{violet}{rgb}{0.5,0.0,1.0}
\definecolor{gaporange}{rgb}{0.776,0.420,0.169}

\title{\LARGE \bf
When Does Touch Matter? Charting the \\
Vision--Interaction Gap in Cluttered Dexterous Grasping\\
}

\author{Hao Jiang$^{1}$, Luis Dominguez$^{2}$, Daniel Seita$^{1}$%
\thanks{$^{1}$Hao Jiang and Daniel Seita are with the Thomas Lord Department of
Computer Science, University of Southern California, USA. Correspondence:
{\tt\small hjiang86@usc.edu}}%
\thanks{$^{2}$Luis Dominguez is with the University of New Mexico, USA.}%
}

\begin{document}
\IEEEaftertitletext{\input{figs/teaser}}
\maketitle
\thispagestyle{empty}
\pagestyle{empty}

\input{sections/abstract}
\input{sections/introduction}
\input{sections/related_work}
\input{sections/problem_statement}
\input{sections/multimodal_policy_design}
\input{sections/experiments}
\input{sections/conclusion}

\renewcommand{\bibfont}{\footnotesize}
\printbibliography

\end{document}

%% file: figs/teaser.tex
\begin{minipage}{\textwidth}
  \centering
  \includegraphics[width=\textwidth]{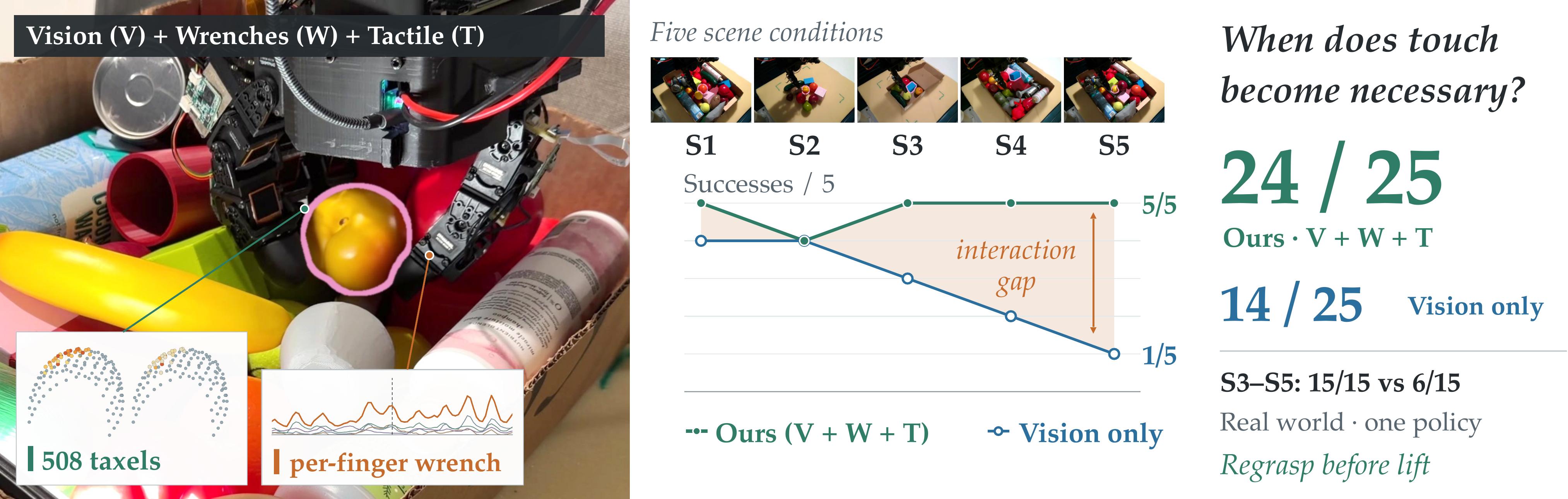}
  \captionof{figure}{Interaction sensing becomes decisive as cluttered dexterous grasping gets harder. Left: one policy grasps a designated target (magenta contour) using RGB\mbox{-}D vision, per-finger wrench estimates, and 508 distributed fingertip taxels (insets). Center: per-condition success over the five scene conditions S1--S5; Vision only falls from 4 / 5 to 1 / 5 while Ours maintains a 4 / 5 to 5 / 5 success rate. The shaded region marks the widening \textit{\textcolor{gaporange}{interaction gap}}. Right: aggregate results, with Ours at 24 / 25 (15 / 15 on S3--S5) against 14 / 25 (6 / 15) for Vision only; the underlying recovery behavior is analyzed in Fig.~\ref{fig:vision_vs_ours}.}
  \label{fig:teaser}
\end{minipage}
\vspace{0.25\baselineskip}

%% file: sections/abstract.tex
\begin{abstract}
Dexterous grasping in clutter poses a basic sensing question: when do tactile measurements and external wrench estimates improve on visual geometry? Occlusion and contact can obscure grasp quality, motivating a controlled evaluation of these interaction signals. We present a controlled real-world study over five tabletop scene conditions on a dexterous hand (LEAP Hand) with vision, per-finger and wrist wrench estimates, and distributed fingertip taxels. With demonstrations, visual observations, action space, and compliant control fixed, we compare diffusion policies trained by imitation learning on vision only, wrench, taxels, or all combined, plus representation and fusion baselines. The combined policy succeeds in 24/25 trials versus 14/25 for vision only, and 15/15 versus 6/15 across the three confined conditions. Ablations show that wrench and taxel feedback are complementary. Behavioral comparisons show that interaction feedback enables earlier rejection of inadequate contacts, regrasping before lift, and more stable grasps. To our knowledge, this is the first real-world study to combine and separately evaluate these interaction modalities for target-oriented dexterous grasping in clutter. These results chart a widening vision--interaction gap and position cluttered dexterous grasping as a benchmark for determining when the learned policy needs interaction sensing. \textit{Project website: {\hypersetup{hidelinks}\href{https://interaction-dex-grasp.github.io/}{interaction-dex-grasp.github.io}}.}
\end{abstract}

%% file: sections/introduction.tex
\section{Introduction}
\label{sec:intro}

Retrieving a specified object from clutter is a fundamental problem in robot manipulation, with applications in warehouse order fulfillment and less structured settings such as retail shelves and homes~\cite{correll2018amazon}. Clutter occludes the target, restricts collision-free approach directions, and creates frequent contact with surrounding objects~\cite{chen2025clutterdexgrasp,zhang2024dexgraspnet}. Although suction cups and parallel-jaw grippers are effective in many bin-picking systems, we focus on the regime in which the robot is equipped with a multi-fingered dexterous hand, such as the LEAP hand~\cite{shaw2023leaphand}. Target retrieval then requires coordinated contacts that singulate and secure the target without destabilizing the scene. Because these demands vary with object geometry and scene clutter, cluttered dexterous grasping provides a benchmark for studying how sensing requirements change with scene and object demands~\cite{fang2020graspnet,zhang2024dexgraspnet}.

Recent work uses visual geometry, including partial point clouds and target masks, to generate dexterous grasps and control closed-loop retrieval in clutter~\cite{chen2025clutterdexgrasp,bai2026retrdex,zhang2024dexgraspnet,zhang2026cadgrasp}. These inputs represent object shape and scene layout, but they do not directly measure load, support, or slip after contact. Visual evidence also degrades when the hand occludes the target or incidental contact changes the scene. Fingertip tactile arrays measure local contact, while external wrench estimates summarize forces and torques at the fingers and wrist. We use \emph{interaction sensing} as an umbrella term for these two signals (tactile and wrench). Our central question is: \emph{when is vision sufficient, and when does interaction sensing become important?}

As summarized in Fig.~\ref{fig:teaser}, we study this question through target-oriented cluttered grasping across five real-world conditions spanning combined scene and object demands. We test the hypothesis that policy-visible interaction sensing has limited effect when visual geometry remains informative and grasp tolerance is high, but becomes important as occlusion, incidental contact, and grasp ambiguity increase.

The physical system combines distributed fingertip tactile sensing with per-finger and wrist external-wrench estimates. Every policy uses the same compliant controller, which holds the robot's contact response constant and isolates the interaction feedback available to the learned policy. All experiments are conducted on the physical system across a controlled suite of tabletop clutter layouts and target objects without condition-specific fine-tuning. We report success across the five scene conditions and use scene states inherited from vision-only failures to compare vision-only and interaction-aware behavior during approaching, regrasping, and lifting stages. The policies perform similarly on easier conditions, but the vision--interaction gap widens with task demand: the combined policy succeeds in 24/25 trials, compared with 14/25 for Vision only. Wrench and taxel ablations show complementary gains, while encoder and fusion variants provide no comparable gain from architectural complexity.

This paper makes the following contributions:

\begin{itemize}
\item \textbf{A demand-conditioned analysis of sensing necessity.} We show how the gap between vision-only and interaction-aware policies changes with scene and object demands in cluttered dexterous grasping.
\item \textbf{An interaction-aware system for cluttered dexterous grasping.} To our knowledge, this is the first real-world system to combine estimated per-finger and wrist wrenches, 508 fingertip taxels, and compliant whole-hand control for target retrieval in clutter, with a shared controller that supports controlled sensing ablations.
\item \textbf{Controlled real-world evaluation against prior methods.} Across 200 physical trials on five tabletop conditions, we analyze closed-loop behavior and separate the effects of sensing modality from encoder and fusion design. Our policy also outperforms strong prior baselines in overall success.
\end{itemize}

%% file: sections/related_work.tex
\section{Related Work}
\label{sec:related}

\subsection{Dexterous Grasping and Manipulation in Clutter}

Classical grasp synthesis formulates contact selection using geometric and mechanical criteria such as wrench-space force closure~\cite{ferrari1992planning}, and mechanics-based planners for clutter deliberately move surrounding objects through prehensile and nonprehensile actions to create feasible grasps~\cite{dogar2011framework}. These methods make contact consequences explicit, but depend on object geometry, friction, and contact models that are difficult to estimate in dense, partially observed scenes.

% Queue the method overview from page 2 so it appears at the top of page 3.
\input{figs/method_overview}

Learning-based grasping reduces this dependence by learning grasp proposals or closed-loop actions from visual observations. Large-scale synthetic and real benchmarks support robust parallel-jaw grasping~\cite{mahler2017dexnet,fang2020graspnet}, while dexterous grasp datasets and generative policies address the larger contact and configuration spaces of multi-fingered hands~\cite{zhang2024dexgraspnet}. Beyond grasp synthesis, visual policies support arm-hand grasping~\cite{DexPoint2022}, multi-stage manipulation~\cite{jiang2026dexmulti}, and sequential multi-object grasping~\cite{lu2026modex,foong2026handful}. Recent systems scale through human video, simulated play, and sim-to-real pretraining~\cite{gupta2026lucid,lum2026play2perfect}. Clutter-specific systems learn object singulation~\cite{Hao2024SopeDex}, target grasping through collision-aware generation, real-world imitation learning, sim-to-real reinforcement learning~\cite{zhang2024dexgraspnet,bai2026retrdex,zhang2026cadgrasp}, and dynamics-aware nonprehensile rearrangement~\cite{zheng2026dapl}. ClutterDexGrasp is the closest vision-only system to our task, and uses a point cloud policy distilled from a clutter-aware reinforcement-learning teacher~\cite{chen2025clutterdexgrasp}. These results establish vision-based cluttered manipulation, but do not characterize how policy-visible contact sensing changes behavior across matched conditions.

\subsection{Contact-Aware Dexterous Manipulation}

Tactile and force sensing expose local contact state that may be ambiguous or occluded in vision~\cite{kappassov2015tactile}. Learned systems use visuo-tactile feedback to predict grasp outcomes, correct poor contacts, and execute closed-loop regrasps~\cite{calandra2018more}. Tactile-only and visuo-tactile policies have since enabled in-hand rotation, contact-rich manipulation, and real-world adaptation~\cite{qi2023rotateit,RobotSynesthesia2024,SeeToTouch2024}. External torque or wrench estimates can improve force-sensitive behavior and object generalization when provided as policy observations~\cite{liu2025factr}. Other approaches use force-aware retargeting or contact-wrench guidance to transfer contact-rich human demonstrations to robots~\cite{wu2026reforce,zhu2026chord}. Contact awareness can also live at the control level: tactile-feedback grasp controllers and proprioceptive contact estimation achieve compliant execution without a learned policy observing touch~\cite{ke2026tacdexgrasp,ma2026current,dou2026neuralactuator}.

In modern learning systems, contact sensing, representation, and fusion are closely coupled. Spatial tactile measurements have been canonicalized in hand-centered coordinates, aligned with kinematics, embedded jointly with vision in 3D, or pretrained through self-supervision, force prediction, and multisensory objectives~\cite{higuera2024sparsh,wu2024canonical,huang2025vitac,huang2025sata,guzey2023dexterity,higuera2025sparshx}. Policies then combine contact with vision and proprioception using feature concatenation, self-attention, cross-attention, or adaptive modality weighting~\cite{chen2023vtt,li2025adaptacdex,heng2025vitacformer}. Other architectures fuse modality-specific diffusion predictions~\cite{chen2025policyconsensus}, separate slow visual planning from fast tactile correction~\cite{xue2025reactive}, steer a visuomotor policy with tactile guidance at inference time~\cite{zhang2026touchguide}, ground actions in generative contact prediction~\cite{xu2026cgp}, or add fast tactile experts to large vision--language--action models~\cite{niu2026trex}. These systems demonstrate the utility of contact feedback, but their hardware, tasks, representations, and policy backbones vary together. Consequently, their performance gains cannot isolate the information supplied by wrench and taxel observations from the architecture used to process them. Our study instead fixes the data, policy backbone, and controller, and separately evaluates sensing content, interaction representation, and fusion across a factor-designed clutter suite.

%% file: figs/method_overview.tex
\begin{figure*}[t]
  \vspace*{5pt}
  \centering
  \includegraphics[width=\textwidth]{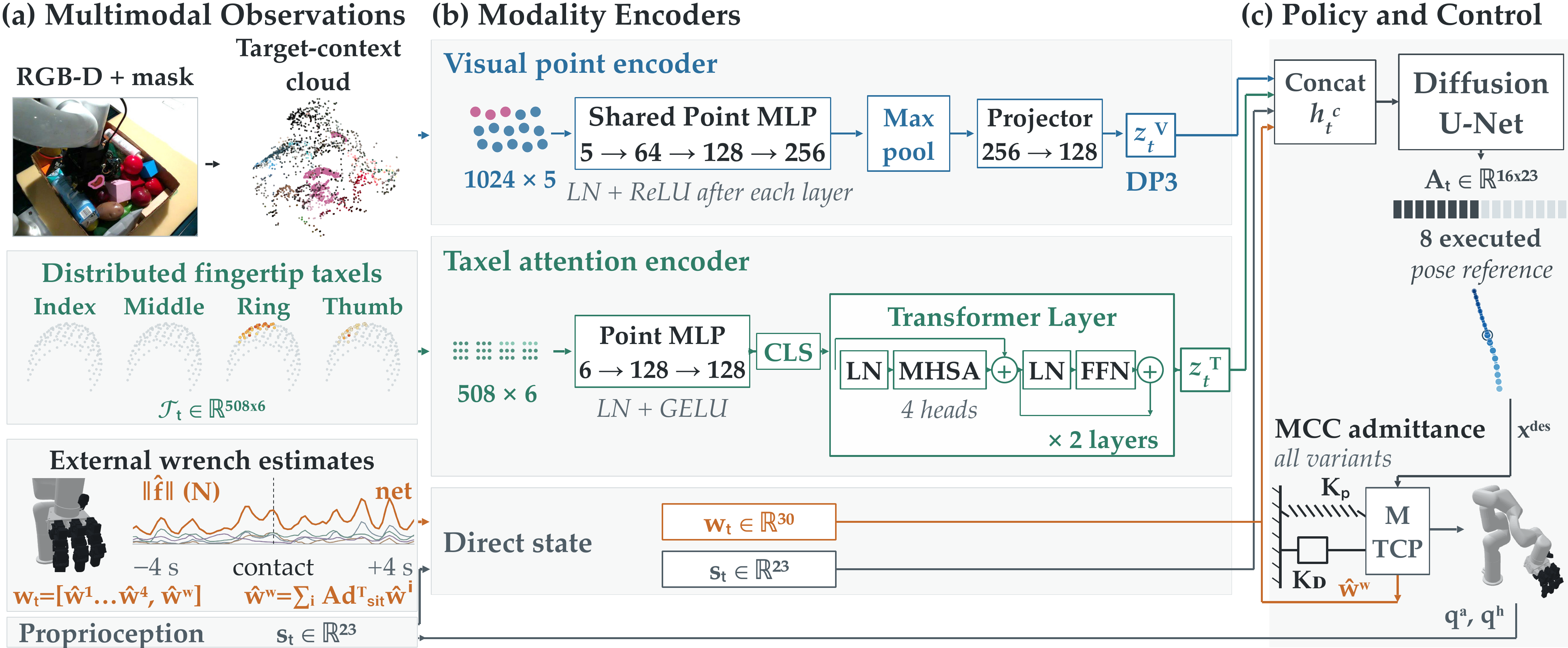}
  \caption{Overview of our multimodal policy and compliant execution stack. (a) An RGB-D target mask is converted into a target--context point cloud; distributed fingertip taxels, external wrench estimates, and proprioception provide complementary interaction and robot-state observations. (b) The visual cloud and taxel field are encoded by a point encoder and a two-layer attention encoder, respectively, while wrench and proprioceptive states are provided directly. (c) The selected representations condition a diffusion policy whose receding-horizon pose and hand-joint references are executed through the shared admittance controller.}
  \label{fig:method_overview}
  \vspace*{-14pt}
\end{figure*}

%% file: sections/problem_statement.tex
\section{Problem Statement and Assumptions}
\label{sec:problem}

We study \emph{target grasping in clutter}. A task condition $d\in\mathcal{E}$ specifies a target object, a set of distractors, and their initial arrangement; success, denoted $Y{=}1$, requires grasping the target, lifting it, and retaining a stable grasp at the end of the episode. Each condition is characterized by scene and object factors $(d_{\rm scene},d_{\rm object})$: the former captures clutter, occlusion, and confinement, while the latter captures target size, geometric regularity, and grasp tolerance. These factors define a composite task spectrum, which Sec.~\ref{ssec:experimental_setup} instantiates as five concrete conditions. At time $t$, the robot has proprioceptive state $\bs_t$ and may receive visual, wrench, and tactile observations $\mathcal{M}_t=\{\mathrm{V}_t,\mathrm{W}_t,\mathrm{T}_t\}$, where $\mathrm{V}_t$ is a scene representation, $\mathrm{W}_t$ contains the estimated local and aggregated external wrenches, and $\mathrm{T}_t$ is a spatially distributed taxel field. The policy outputs the reference action $\ba_t=(\mathbf{x}^{\rm ee}_t,\mathbf{q}^{\rm hand}_t)\in SE(3)\times\mathbb{R}^{d_h}$ for the end-effector pose and hand configuration, where $d_h$ is the hand's degrees of freedom.

Given expert demonstrations $\mathcal{D}=\{\xi_n\}_{n=1}^{N}$, we learn a policy $\pi_{\theta}^{c}$ for each sensing configuration $c\subseteq\{\mathrm{V},\mathrm{W},\mathrm{T}\}$, with observation $\bo_t^{c}=(\bs_t,\{m_t:m\in c\})$. We evaluate each policy by $J_c(d)=\Pr_{\pi_{\theta}^{c}}(Y{=}1\mid d)$ and study how the vision--interaction gap $\Delta(d)=J_{\mathrm{VWT}}(d)-J_{\mathrm{V}}(d)$ changes across the task spectrum, and whether $\mathrm{W}$ and $\mathrm{T}$ provide complementary information. We assume synchronized and spatially calibrated observations, a specified target and common success criterion, but no ground-truth object pose or contact state; tactile sensing covers every fingertip. Across comparisons, the demonstrations, action interface, low-level compliant controller, and training protocol are fixed; only the policy-visible modalities and their representation or fusion vary.

%% file: sections/multimodal_policy_design.tex
\section{Multimodal Policy Design}
\label{sec:method}

Figure~\ref{fig:method_overview} instantiates the sensing configurations from Sec.~\ref{sec:problem}. Learned encoders map the visual and taxel observations to compact features, while normalized proprioceptive and wrench states are provided directly. The selected inputs condition a policy that generates nominal arm and hand references for compliant execution.

\subsection{Visual and Proprioceptive Representations}

Given the target name, we detect and track its image masks using Grounding DINO~\cite{liu2023grounding} and SAM~2~\cite{ravi2024sam}. Registered depth pixels from the calibrated views are back-projected and fused in the world frame. We preserve both the target and a local crop of the surrounding scene, represented as
\begin{equation}
\mathbf{P}_t=\{(\mathbf{p}_{t,j},\ell_{t,j},d_{t,j})\}_{j=1}^{N_v}
\in\mathbb{R}^{N_v\times5},
\label{eq:visual-cloud}
\end{equation}
where $\mathbf p_{t,j}\in\mathbb R^3$ is a world-frame point, $\ell_{t,j}\in\{0,1\}$ indicates context or target, and $d_{t,j}$ is its distance to the target centroid. We stratify the cloud to $N_v{=}1024$ points while preserving the target-to-context ratio. A DP3-style point encoder, comprising a shared pointwise MLP, global max pooling, and linear projection, maps $\mathbf P_t$ to $\mathbf z_t^{\rm V}\in\mathbb R^{128}$~\cite{Ze2024DP3}.

% Queue the single-column setup late on page 3 so it appears at the top of
% page 4.
\afterpage{\input{figs/real_world_setup}}

The proprioceptive state contains the measured end-effector position and quaternion with the hand joint positions:
\begin{equation}
\mathbf{s}_t=(\mathbf p_t^{\rm ee},\mathbf q_t^{\rm ee},
\boldsymbol\theta_t^{\rm hand})\in\mathbb R^{23}.
\label{eq:proprio-state}
\end{equation}
After channel-wise normalization, $\mathbf s_t$ is passed directly to the policy without a separate learned encoder.

\subsection{Interaction-Sensing Representations}
\label{ssec:interaction_sensing}

The LEAP hand provides 16 actuated DoF across its thumb, index, middle, and ring fingers in a low-cost robot-learning platform~\cite{shaw2023leaphand}. We mount one PaXini pad on each fingertip, together providing $N_\tau{=}508$ spatially distributed three-axis taxels. This pairing combines multi-finger actuation with spatially resolved contact measurements and has been used for real-world contact-rich dexterous tasks~\cite{wu2024canonical,li2025adaptacdex}. Taxel $j$ is represented by its position and measured three-axis force in the world frame, forming the force-annotated point set
\begin{equation}
\mathcal T_t=\{(\mathbf p_{t,j}^{\tau},\mathbf f_{t,j}^{\tau})\}_{j=1}^{N_\tau}
\in\mathbb R^{508\times6}.
\label{eq:taxel-observation}
\end{equation}
For each taxel, let $\mathbf x_{t,j}^{\tau}=(\mathbf p_{t,j}^{\tau}, \mathbf f_{t,j}^{\tau})\in\mathbb R^6$. Our encoder is
\begin{equation}
\begin{aligned}
\mathbf E_t^{(0)}
  &= [\mathbf e_{\rm cls};\psi_\tau(\mathbf x_{t,1}^{\tau});\ldots;
      \psi_\tau(\mathbf x_{t,N_\tau}^{\tau})], \\
\mathbf E_t^{(\ell+1)}
  &= \operatorname{Block}_{\ell}(\mathbf E_t^{(\ell)}),\quad \ell=0,1, \\
\mathbf z_t^{\rm T}
  &= \mathbf W_\tau\mathbf E_{t,0}^{(2)}\in\mathbb R^{128},
\end{aligned}
\label{eq:taxel-attention}
\end{equation}
where $\psi_\tau$ is a shared pointwise MLP, $\mathbf e_{\rm cls}$ is a learned class token, and each $\operatorname{Block}_{\ell}$ is a four-head self-attention transformer block. Because position is part of every taxel token, the class-token readout summarizes both force and contact geometry without prescribing individual finger contributions.

Following Minimalist Compliance Control (MCC)~\cite{shi2026minimalist}, we also estimate each fingertip wrench $\hat{\mathbf{w}}^{i}=(\hat{\mathbf{f}}^{i},\hat{\mathbf{m}}^{i})\in\mathbb{R}^{6}$ from actuator currents and the hand Jacobian, expressed in fingertip frame $S_i$. We transform the four estimates into the wrist tool frame $T$ and sum them for arm-level feedback:
\begin{equation}
\hat{\mathbf{w}}^{\rm wrist}
=\sum_{i=1}^{4}\mathrm{Ad}_{S_iT}^{\top}\hat{\mathbf{w}}^{i}.
\label{eq:wrench-aggregation}
\end{equation}
Here $\mathrm{Ad}_{S_iT}^{\top}$ denotes the dual-adjoint wrench transformation from $S_i$ to $T$, including the moment induced by the frame offset. These estimates form the policy wrench observation
\begin{equation}
\mathbf w_t=(\hat{\mathbf w}_t^1,\ldots,\hat{\mathbf w}_t^4,
\hat{\mathbf w}_t^{\rm wrist})\in\mathbb R^{30}.
\label{eq:wrench-observation}
\end{equation}
After channel-wise normalization, $\mathbf w_t$ is provided directly to policies for which $\mathrm W\in c$.

A graph neural network (GNN) tests whether explicit local contact relationships improve on taxel attention, using force-aware message passing over a $k$-nearest-neighbor taxel graph. FC-full is a joint taxel--wrench encoder: it hierarchically combines per-pad taxel features with net pad forces, per-finger external wrenches and poses, and the aggregated wrist wrench. Thus, FC-full changes how the two interaction modalities are encoded and fused, whereas the primary policy encodes taxels with~\eqref{eq:taxel-attention} and appends $\mathbf w_t$ directly.

% This point is reached on page 4, so the full-width task suite appears at the
% top of page 5 immediately above the experiments.
\input{figs/difficulty_suite}

\subsection{Multimodal Fusion and Action Generation}

Diffusion policies are effective action generators for visuomotor control, 3D manipulation, and high-dimensional dexterous hand tasks including grasping in clutter~\cite{chi2023diffusionpolicy,Ze2024DP3,chen2025clutterdexgrasp}. We fix this backbone across sensing ablations. Our primary policy uses feature-level concatenation. For sensing configuration $c$, we form
\begin{equation}
\mathbf h_t^c=\mathbf z_t^{\rm V}\oplus\mathbf s_t
\oplus[\mathbf w_t]_{\mathrm W\in c}
\oplus[\mathbf z_t^{\rm T}]_{\mathrm T\in c},
\label{eq:feature-fusion}
\end{equation}
where $\oplus$ denotes concatenation and a bracketed term is omitted when its channel is unavailable. Features from the latest two observations condition a single diffusion denoiser. It predicts a 16-step sequence of 23-D end-effector pose and hand-joint references, of which the first eight are executed before replanning. We refer to the full $\mathrm{VWT}$ configuration with taxel attention and direct concatenation as \emph{VWT-Attn (Ours)} in the experiments. Table~\ref{tab:real_robot_success} lists the policy-visible inputs for every comparison.

To separate information content from fusion design, we additionally implement two structured fusion baselines over the same channels. First, PolicyConsensus trains modality-specific diffusion denoisers and combines their predictions using learned consensus weights~\cite{chen2025policyconsensus}. Second, Reactive Diffusion Policy (RDP) samples a slow latent action plan and repeatedly decodes it using current wrench and taxel observations for fast within-chunk correction~\cite{xue2025reactive}.

\subsection{Compliant Execution}
\label{ssec:compliant_execution}

Because contact with surrounding objects is unavoidable in clutter, every policy executes its nominal arm and hand position references through the same task-space compliant controller. This accommodates incidental contact and fixes the low-level response across sensing ablations. Whether or not the learned policy observes $\mathrm W$, the estimated wrenches enter an admittance law applied separately to the arm and fingertips. Sensing ablations therefore change only what the policy observes: even Vision only executes through this wrench-based compliance. For task-space state $\mathbf{x}$ and nominal policy reference $\mathbf{x}_{\rm des}$,
\begin{equation}
\mathbf{M}\ddot{\mathbf{x}}
=\mathbf{K}_p(\mathbf{x}_{\rm des}-\mathbf{x})
+\mathbf{K}_d(\dot{\mathbf{x}}_{\rm des}-\dot{\mathbf{x}})
+\mathbf{u}_{\rm ext}.
\label{eq:admittance-control}
\end{equation}
Here $\mathbf{M}$, $\mathbf{K}_p$, and $\mathbf{K}_d$ are the virtual inertia, stiffness, and damping matrices. We use $\mathbf{u}_{\rm ext}=\hat{\mathbf{f}}^{i}$ for fingertip translation and $\mathbf{u}_{\rm ext}=\hat{\mathbf{w}}^{\rm wrist}$ for arm motion. Contact can therefore displace the executed motion from the nominal policy reference. Taxel measurements do not enter the low-level controller.

%% file: figs/real_world_setup.tex
\begin{figure}[t]
  \vspace*{5pt}
  \centering
  \includegraphics[width=\columnwidth]{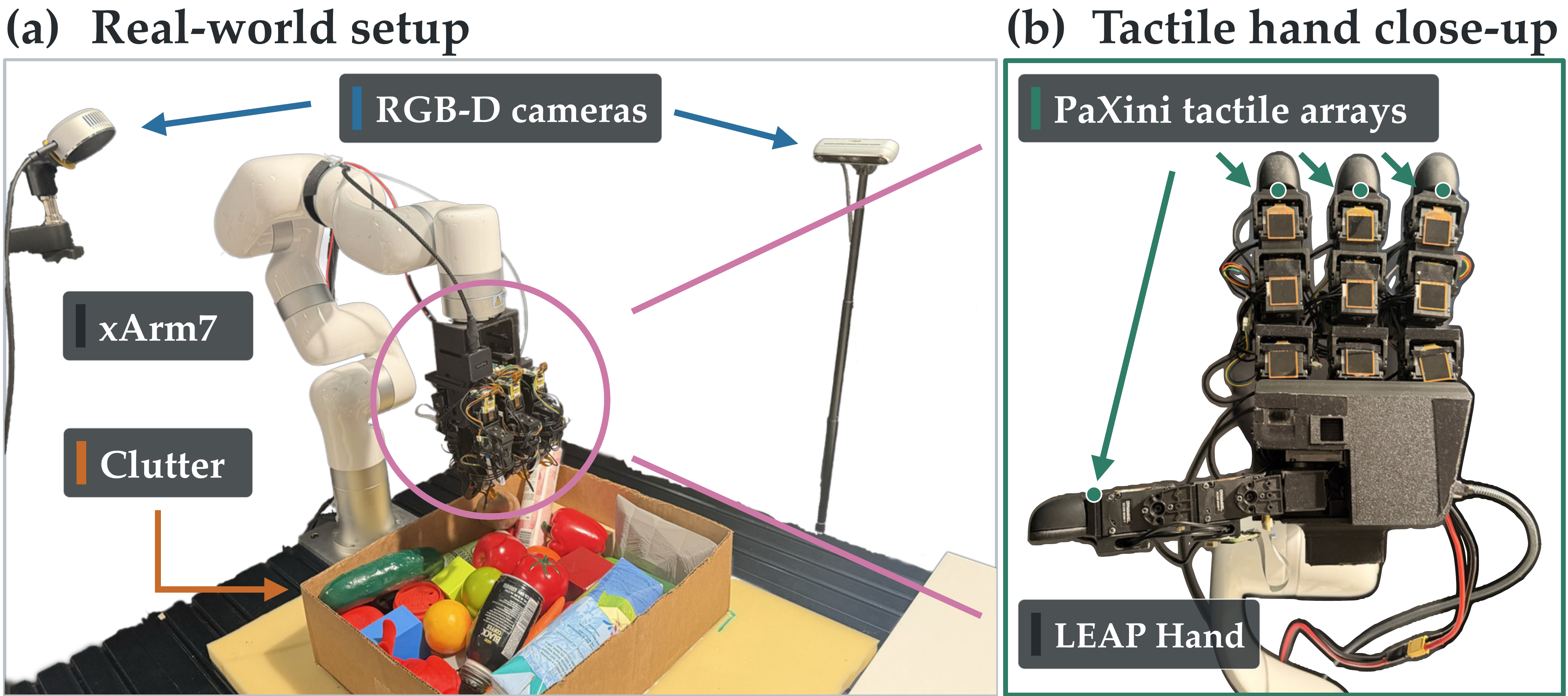}
  \caption{Real-world platform. (a) The xArm7--LEAP Hand system operates in clutter under two calibrated RGB\mbox{-}D views. (b) Four PaXini fingertip arrays provide 508 three-axis taxels.}
  \label{fig:real_world_setup}
  \vspace*{-14pt}
\end{figure}

%% file: figs/difficulty_suite.tex
\begin{figure*}[t]
  \vspace*{5pt}
  \centering
  \includegraphics[width=\textwidth]{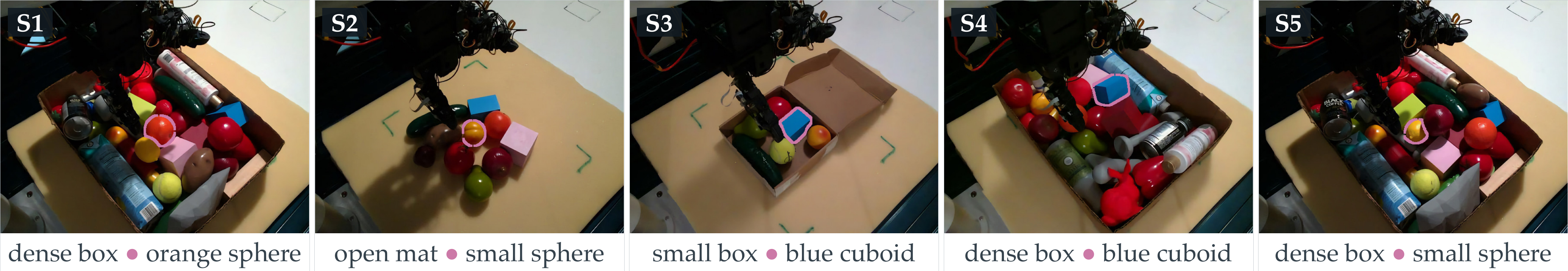}
  \caption{The five real-world scene conditions. Labels state each condition\textquotesingle s scene and object factors, fixed before evaluation; magenta contours identify the targets. S1 and S2 dissociate scene and object challenges (heavy clutter with a grasp-tolerant target; a grasp-sensitive target in the open), while S3--S5 form a nested chain that adds confinement, then surrounding clutter, then a small rim-packed target.}
  \label{fig:difficulty_suite}
  \vspace*{-14pt}
\end{figure*}

%% file: sections/experiments.tex
\section{Experiments}
\label{sec:experiments}

Our experiments instantiate the sensing configurations and task conditions of Sec.~\ref{sec:problem} to answer four questions: (1) when interaction sensing becomes important as a policy observation, (2) how interaction feedback changes closed-loop behavior after contact, (3) whether wrench and taxel observations are complementary, and (4) whether more structured representations or fusion improve performance when the policy receives the same information. Secs.~\ref{ssec:interaction_gap}--\ref{ssec:architecture_ablation} answer these questions using the shared real-world evidence of Table~\ref{tab:real_robot_success}, and Sec.~\ref{ssec:failure_modes} analyzes failure modes.

\subsection{Experimental Setup}
\label{ssec:experimental_setup}

\paragraph{Robot Platform and Control} Figure~\ref{fig:real_world_setup} shows the physical platform. We use a 7-DoF xArm7 with the LEAP--PaXini tactile hand (Sec.~\ref{ssec:interaction_sensing}); calibrated Intel RealSense L515 and D455 cameras provide two RGB\mbox{-}D views. At 10\,Hz, the policy commands an end-effector pose and 16 hand joint positions through the controller of Sec.~\ref{ssec:compliant_execution}. The controller and its wrench feedback are fixed across controlled policies, so removing $\mathrm W$ hides wrench estimates from the learned policy without disabling physical compliance.

\paragraph{Demonstrations and Training} We collect 320 expert demonstrations over three rounds using a Vive tracker for arm teleoperation and a Rokoko glove for hand teleoperation. Each demonstration records synchronized robot state, calibrated multi-view depth, per-finger and wrist wrench estimates, and four fingertip taxel arrays. Each controlled policy is trained from scratch on the same demonstrations with world-frame translation augmentation, a 90/10 training--validation split, 300 epochs, and the horizons of Sec.~\ref{sec:method}.

\paragraph{Policy Configurations} Table~\ref{tab:real_robot_success} reports our main results. % makes the policy-visible inputs and controlled comparisons explicit.
Vision only, Wrench, Taxel-Attn, and VWT-Attn isolate sensing; GNN and FC-full change the interaction representation; and PolicyConsensus and RDP change fusion or action generation. We additionally test ClutterDexGrasp~\cite{chen2025clutterdexgrasp} as an external vision-only, reinforcement learning method.

\paragraph{Scene Conditions} Figure~\ref{fig:difficulty_suite} shows the five conditions, each specified before any evaluation by its scene and object factors. S1 is densely cluttered, but its large exposed sphere is grasp tolerant; S2 places a small, grasp-sensitive sphere on an open mat; S3 introduces box confinement but uses a regular cuboid; S4 surrounds that cuboid with dense box clutter; and S5 is rim packed with a small round target. The suite is structured by design: S3--S5 form a nested chain in which each condition adds one challenge to the previous one (surrounding clutter in S4, then a grasp-sensitive rim-packed target in S5), while S1 and S2 dissociate scene and object challenges by pairing heavy clutter with a tolerant target and an open scene with a sensitive one.

\paragraph{Evaluation Protocol} We conduct five physical trials per policy--condition combination and reset the scene between trials. Resets are performed by hand against a per-condition reference layout, so object placement varies slightly across trials. Each trial is an episode that permits repeated autonomous grasp attempts; success is assessed over the episode. Success requires grasping the specified target, lifting it, and retaining a stable grasp at the episode end without human intervention; all other outcomes count as failures. Hardware, perception, action rate, and stopping criteria are fixed. Eight controlled variants across five conditions yield $8\times5\times5=200$ trials. To promote transparency, we will release videos of all evaluation trials on our project website.
% We report raw success counts because each policy--condition combination contains only five trials.

\subsection{When Does Interaction Sensing Matter?}
\label{ssec:interaction_gap}

Ours succeeds in 24 of 25 trials, compared with 14 of 25 for Vision only (Table~\ref{tab:real_robot_success}). More important than this 40 percentage-point aggregate difference is how it distributes across conditions. On the two dissociated conditions the gap is small or absent: one success out of five on the cluttered but grasp-tolerant S1, and a 4-of-5 tie on the open-scene S2. Along the nested chain S3--S5 it widens monotonically from two to four successes out of five: Ours succeeds in every trial, while Vision only declines from 3 of 5 to 1 of 5 as clutter and then a grasp-sensitive target are added. Interaction information thus has limited measurable value while the grasp remains observable and forgiving, but becomes consequential as these challenges compound.

% Queue the two results tables and recovery figure together for the next page.
\input{tabs/real_robot_success}
\input{tabs/execution_efficiency}
\input{figs/vision_vs_ours}

ClutterDexGrasp provides an external vision-only reference point: a clutter-aware RL teacher distilled into a point cloud diffusion policy for zero-shot sim-to-real deployment~\cite{chen2025clutterdexgrasp}. We reproduce the student using DP3 with the 5-D target--context point-cloud representation in Eq.~\eqref{eq:visual-cloud}, train it in simulation, and deploy it zero-shot on hardware. Following the original implementation, we command absolute hand joint targets without our admittance controller. The reproduced student achieves 50.2\% success in simulation. On hardware, however, it exhibited unsafe rapid motions and continued pushing downward against the table after contact, requiring an emergency stop.

\subsection{How Interaction Feedback Changes Behavior}
\label{ssec:behavioral_analysis}

Vision only has no direct measure of grasp quality before lifting: it often discovers an inadequate grasp only through a failed lift and repeats a lift--fail--regrasp cycle, and on small, heavily occluded targets the visible point cloud becomes too sparse to indicate finger support. Ours instead detects weak or imbalanced contact from the wrench and taxel observations and initiates regrasping before lifting.

Table~\ref{tab:execution_efficiency} quantifies this behavior among successful episodes. Ours has a lower mean duration in every condition, a lower median in four of five (on S3 the three successful Vision-only episodes have a lower median), and fewer full lift attempts throughout, consistent with regrasping before a committed lift. Because these metrics condition on success, they carry survivorship bias and must be read jointly with Table~\ref{tab:real_robot_success}, especially for Vision only on S4 and S5 (two and one successful episodes).

\noindent\textbf{Recovery case study.} Figure~\ref{fig:vision_vs_ours} localizes the behavioral gap in a scene inherited from a failed Vision-only rollout: Vision only repeatedly makes contact but lifts without the target, and its terminal state then initializes Ours, which rejects the inadequate contact, regrasps, stabilizes the target, and completes the lift. Since Ours must recover in a scene Vision only has already disturbed, the demonstration is conservative rather than favorable to Ours.

\noindent\textbf{Robustness.} Figure~\ref{fig:qualitative_robustness} shows three distinct behaviors of Ours: persistence through four unsuccessful attempts under dense clutter until the target becomes accessible, taxel-visible refinement from initial contact to a supported three-finger enclosure before lifting, and recovery from two external perturbations of the target.

\input{figs/qualitative_robustness}
% Queue the failure cases with the robustness figure on the following page.
\input{figs/failure_cases}

\subsection{Wrench--Taxel Complementarity}
\label{ssec:modality_ablation}

Table~\ref{tab:real_robot_success} compares the single-interaction variants against their combination. 
% Wrench only achieves 15/25 and taxel only 18/25, compared with 24/25 for Ours (and 14/25 for Vision only). 
The advantage is concentrated along the chain S3--S5: Ours achieves 5/5 throughout, while the stronger single-interaction results are 3/5, 4/5, and 3/5, respectively.

The two signals resolve different decisions. Wrench feedback exposes weak or obstructed loading early, prompting the policy to abandon an unsuccessful attempt and regrasp before lifting. The dense taxel field instead resolves how contact is distributed across the fingertips, distinguishing incidental contact from a supported multi-finger grasp and indicating when the object is ready to lift. Their combination supplies an early regrasp cue and a grasp-quality cue, explaining why neither modality alone recovers the full gain.

\subsection{Interaction Representation and Fusion Ablations}
\label{ssec:architecture_ablation}

Table~\ref{tab:real_robot_success} summarizes the representation and fusion comparison. 
% Ours achieves 24/25, compared with 21/25 for GNN, 21/25 for RDP, 17/25 for FC-full, and 15/25 for PolicyConsensus. 
GNN and RDP are the strongest alternatives, but neither matches the overall success of direct taxel-attention fusion. FC-full and PolicyConsensus achieve substantially lower success despite receiving the same sensing inputs, highlighting the importance of how interaction information is represented and fused.

\noindent\textbf{Interaction representation.} The GNN produces active corrections but over-reacts after contact, occasionally releasing an established grasp: relational message passing appears prone to fitting sample-specific contact patterns at our data scale, whereas the attention class token forms a global grasp summary and commits more decisively. FC-full performs worse despite its hand-designed hierarchy, suggesting that premature compression makes the relevant contact criterion harder to learn. Both results favor a simple global interaction summary for the available data. % ; whether the ranking changes at larger data scale remains open.

\noindent\textbf{Fusion and action generation.} PolicyConsensus~\cite{chen2025policyconsensus} factorizes modalities into separate diffusion experts under a learned router, which increases inference latency and collapses to a static, vision-dominated gate: logged over a full deployment, it assigns approximately $0.95$ weight to the vision expert and varies by less than $0.4$ percentage points across approach, contact, and lift. RDP~\cite{xue2025reactive} performs comparably to GNN and succeeds in all five S5 trials, but offers no aggregate advantage over direct fusion. The shared compliant controller may reduce the benefit of its fast contact corrections. We use sensor-local taxel inputs; world-frame coordinates expose end-effector pose and produced an arm-stationary policy.

\subsection{Failure Modes}
\label{ssec:failure_modes}

Figure~\ref{fig:failure_cases} identifies two policy limitations. First, when the target becomes fully occluded in both policy RGB\mbox{-}D views, the policy cannot maintain or recover a sufficiently accurate target-state estimate. Second, after reaching the target region, the policy can enter a local action stall with no meaningful progress. % for 35~s.
These failures expose two missing capabilities: persistent object state estimation across occlusion and explicit progress monitoring with a recovery mechanism. % The external camera used for visualization is not an input to the policy.

% \FloatBarrier

%% file: tabs/real_robot_success.tex
\begin{table*}[t]
  \vspace*{5pt}
  \centering
  \caption{Real-world target-grasp success across the five scene conditions. The Input column lists policy-visible modalities; all controlled variants also receive the same proprioceptive state $\mathbf s_t$ and use the shared compliant controller. Results are reported as \textbf{successes / total trials}. The full VWT-Attn configuration is highlighted in bold; tied baselines are not additionally bolded.}
  \label{tab:real_robot_success}
  \vspace*{-5pt}
  \small
  \setlength{\tabcolsep}{4pt}
  \begin{tabular}{@{}llcccccc@{}}
    \toprule
    \multirow{2}{*}{Method} &
    \multirow{2}{*}{Input} &
    \multicolumn{5}{c}{Scene Condition} &
    \multirow{2}{*}{Overall} \\
    \cmidrule(lr){3-7}
    & & S1 & S2 & S3 & S4 & S5 & \\
    \midrule
    Vision only
    & $\mathrm V$ & 4 / 5 & 4 / 5 & 3 / 5 & 2 / 5 & 1 / 5 & 14 / 25 (56\%) \\
    Wrench
    & $\mathrm{VW}$ & 5 / 5 & 3 / 5 & 3 / 5 & 3 / 5 & 1 / 5 & 15 / 25 (60\%) \\
    Taxel-Attn
    & $\mathrm{VT}$ & 5 / 5 & 4 / 5 & 2 / 5 & 4 / 5 & 3 / 5 & 18 / 25 (72\%) \\
    \cmidrule(lr){1-8}
    GNN
    & $\mathrm{VWT}$ & 5 / 5 & 4 / 5 & 4 / 5 & 4 / 5 & 4 / 5 & 21 / 25 (84\%) \\
    FC-full
    & $\mathrm{VWT}$ & 5 / 5 & 3 / 5 & 5 / 5 & 3 / 5 & 1 / 5 & 17 / 25 (68\%) \\
    PolicyConsensus
    & $\mathrm{VWT}$ & 4 / 5 & 3 / 5 & 5 / 5 & 2 / 5 & 1 / 5 & 15 / 25 (60\%) \\
    Reactive DP
    & $\mathrm{VWT}$ & 4 / 5 & 4 / 5 & 4 / 5 & 4 / 5 & 5 / 5 & 21 / 25 (84\%) \\
    \midrule
    \textbf{VWT-Attn (Ours)}
    & $\mathrm{VWT}$ & \textbf{5 / 5} & \textbf{4 / 5} & \textbf{5 / 5} & \textbf{5 / 5} & \textbf{5 / 5}
    & \textbf{24 / 25 (96\%)} \\
    \bottomrule
  \end{tabular}
  \vspace*{-5pt}
\end{table*}

%% file: tabs/execution_efficiency.tex
\begin{table*}[!t]
  \centering
  \caption{Execution efficiency among successful episodes across the five scene conditions. Duration is measured from the first policy action until the measured EEF remains at least 10~cm above its pre-final-lift height for 0.3~s; each entry reports \textbf{median / mean} seconds. A full lift attempt is a post-grasp upward EEF excursion of at least 10~cm; smaller excursions are treated as regrasp motions. Lift entries report the mean. Lower is better for both metrics ($\downarrow$); success rates are reported separately in Table~\ref{tab:real_robot_success}.}
  \label{tab:execution_efficiency}
  \small
  \setlength{\tabcolsep}{9pt}
  \begin{tabular}{@{}llccccc@{}}
    \toprule
    Metric & Method & S1 & S2 & S3 & S4 & S5 \\
    \midrule
    \multirow{2}{*}{Duration (s) $\downarrow$}
    & Vision only & 15.0 / 19.0 & 18.1 / 18.5 & \textbf{14.5} / 22.7 & 38.8 / 38.8 & 41.6 / 41.6 \\
    & \textbf{VWT-Attn (Ours)} & \textbf{9.8 / 14.1} & \textbf{14.9 / 15.7} & 25.3 / \textbf{22.5} & \textbf{13.9 / 24.7} & \textbf{30.8 / 39.4} \\
    \cmidrule(lr){1-7}
    \multirow{2}{*}{Full lift attempts $\downarrow$}
    & Vision only & 1.50 & 1.50 & 1.67 & 2.50 & 3.00 \\
    & \textbf{VWT-Attn (Ours)} & \textbf{1.00} & \textbf{1.00} & \textbf{1.20} & \textbf{2.00} & \textbf{2.40} \\
    \bottomrule
  \end{tabular}
  \vspace*{-5pt}
\end{table*}

%% file: figs/vision_vs_ours.tex
\begin{figure*}[!t]
  \centering
  \includegraphics[width=\textwidth]{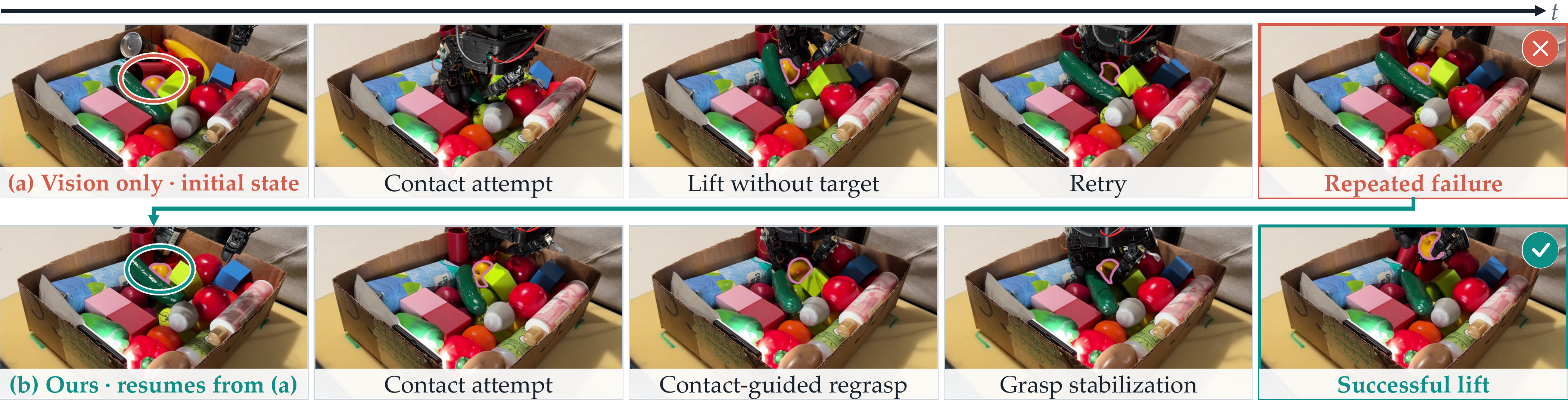}
  \caption{Vision-only failure and interaction-aware recovery. (a) Vision only repeatedly contacts the target region but lifts without the target. (b) Starting from the terminal state of (a), Ours uses interaction feedback to regrasp, stabilize the contact, and successfully lift the target.}
  \label{fig:vision_vs_ours}
  \vspace*{-14pt}
\end{figure*}

%% file: figs/qualitative_robustness.tex
\begin{figure*}[!t]
  \vspace*{5pt}
  \centering
  \includegraphics[width=\textwidth]{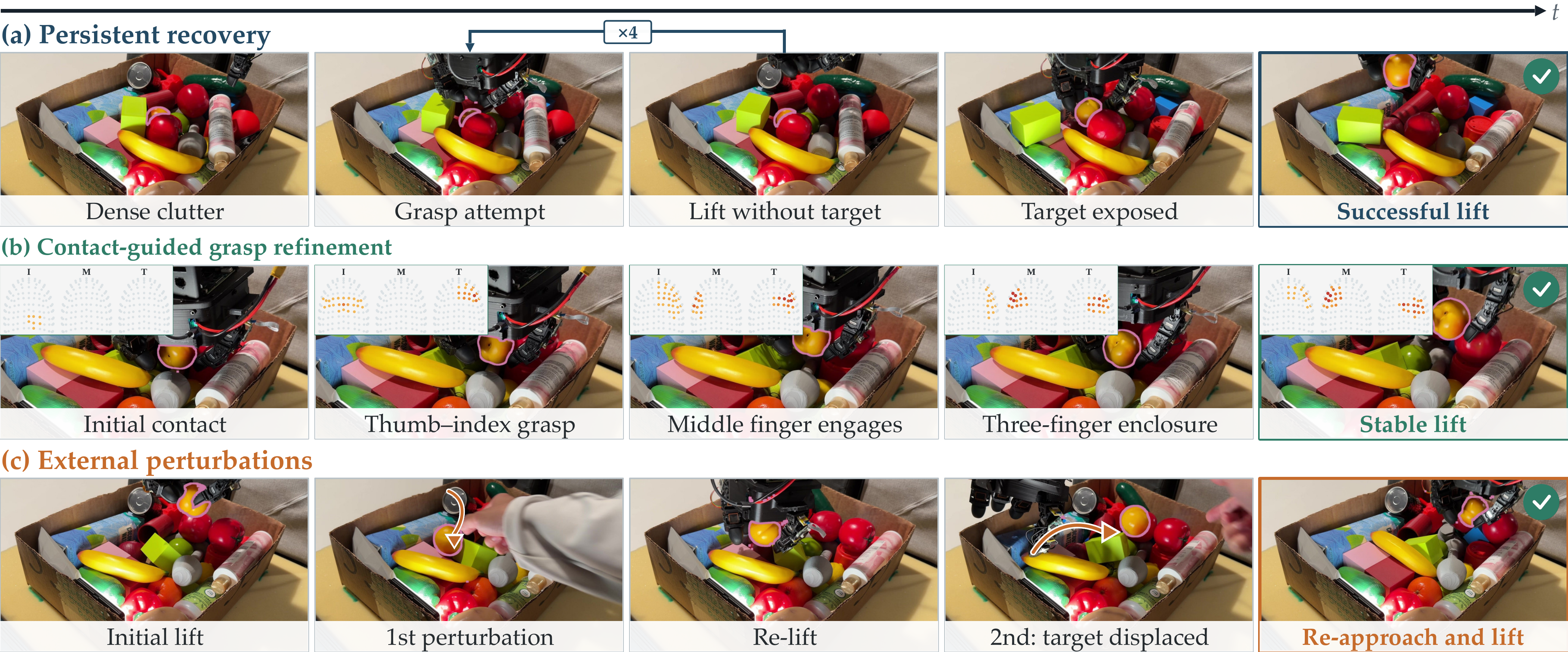}
  \caption{Qualitative performance and robustness of Ours. (a) Persistent recovery under dense clutter exposes the target and enables a subsequent successful lift. (b) Contact-guided refinement progresses from initial contact to a three-finger enclosure and stable lift; I, M, and T denote the index, middle, and thumb taxel arrays, and warmer colors indicate larger force magnitude. (c) The policy recovers from two external perturbations, re-approaching the displaced target and completing the lift.}
  \label{fig:qualitative_robustness}
  \vspace*{-5pt}
\end{figure*}

%% file: figs/failure_cases.tex
\begin{figure*}[!t]
  \centering
  \includegraphics[width=\textwidth]{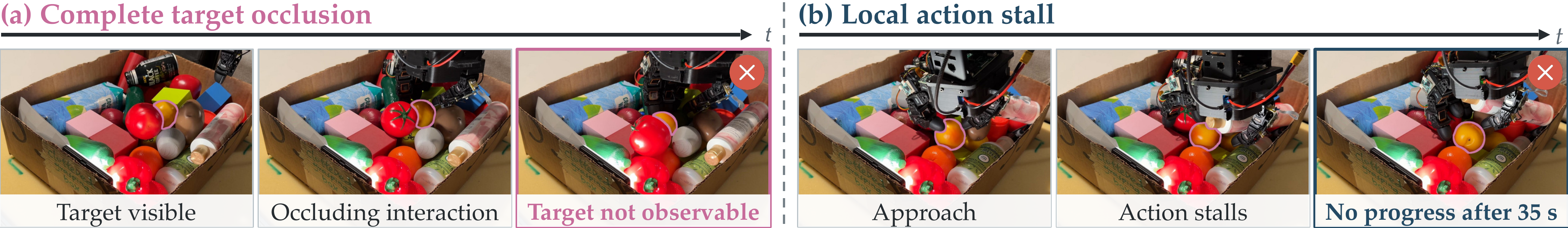}
  \caption{Representative failure cases of Ours, shown from an external camera that is not observed by the policy. (a) The target becomes simultaneously occluded in both policy RGB\mbox{-}D views during interaction, preventing reliable relocalization. (b) The policy approaches the target region but enters a local action stall, producing no meaningful progress over 35~s.}
  \label{fig:failure_cases}
  \vspace*{-14pt}
\end{figure*}

%% file: sections/conclusion.tex
\section{Conclusion, Limitations, and Future Work}
\label{sec:conclusion}

\noindent\textbf{Conclusion.} We present a controlled real-world study of when interaction sensing becomes beneficial for learned dexterous grasping in clutter. Holding everything but the policy-visible sensing fixed, we find that the vision--interaction gap widens with task demand, that wrench and taxel feedback supply complementary cues, and that simple direct fusion outperforms more structured representation and fusion alternatives. As occlusion and contact ambiguity grow, our results suggest treating interaction sensing as a first-class policy observation rather than an optional refinement.

\noindent\textbf{Limitations.} We evaluate using a single embodiment, the LEAP hand, which bounds attainable success: its thick fingertips make it difficult to squeeze into narrow gaps and singulate the target from close neighbors. Our interaction sensing also covers one hardware class; other force modalities, such as vision-based tactile sensors, remain untested.

\noindent\textbf{Future work.} Future work could extend this study to a larger-scale benchmark spanning multiple embodiments and tactile sensor categories, explore improved dexterous mechanical designs that ease singulation in tight clutter, and develop persistent state estimation that blends visual and tactile evidence for robustness under occlusion.

\section*{Acknowledgments}
\label{sec:acknowledgments}

We acknowledge generous hardware support from PaXini. Daniel Seita and Luis
Dominguez are supported in part by NSF grant \#2447397.

%% file: example.bib
@String { icra    = {IEEE International Conference on Robotics and Automation (ICRA)} }

@String { iros    = {IEEE/RSJ International Conference on Intelligent Robots and Systems (IROS)} }

@String { isrr    = {International Symposium on Robotics Research (ISRR)} }

@String { rss     = {Robotics: Science and Systems (RSS)} }

@String { neurips = {Neural Information Processing Systems (NeurIPS)} }

@String { iclr    = {International Conference on Learning Representations (ICLR)} }

@String { corl    = {Conference on Robot Learning (CoRL)} }

@String { cvpr    = {IEEE/CVF Conference on Computer Vision and Pattern Recognition (CVPR)} }

@String { eccv    = {European Conference on Computer Vision (ECCV)} }

@String { roman   = {IEEE International Conference on Robot and Human Interactive Communication (RO-MAN)} }

@inproceedings{RobotSynesthesia2024,
  title={{Robot Synesthesia: In-Hand Manipulation with Visuotactile Sensing}},
  author={Ying Yuan and Haichuan Che and Yuzhe Qin and Binghao Huang and Zhao-Heng Yin and Kang-Won Lee and Yi Wu and Soo-Chul Lim and Xiaolong Wang},
  booktitle = icra,
  year={2024},
}

@inproceedings{SeeToTouch2024,
  title={{See to Touch: Learning Tactile Dexterity through Visual Incentives}},
  author={Irmak Guzey and Yinlong Dai and Ben Evans and Soumith Chintala and Lerrel Pinto},
  booktitle=icra,
  year={2024},
}

@inproceedings{zhang2024dexgraspnet,
  title={{DexGraspNet 2.0: Learning Generative Dexterous Grasping in Large-scale Synthetic Cluttered Scenes}},
  author={Zhang, Jialiang and Liu, Haoran and Li, Danshi and Yu, XinQiang and Geng, Haoran and Ding, Yufei and Chen, Jiayi and Wang, He},
  booktitle=corl,
  year={2024}
}

@inproceedings{chen2025clutterdexgrasp,
  title={{ClutterDexGrasp: A Sim-to-Real System for General Dexterous Grasping in Cluttered Scenes}},
  author={Chen, Zeyuan and Yan, Qiyang and Chen, Yuanpei and Wu, Tianhao and Zhang, Jiyao and Ding, Zihan and Li, Jinzhou and Yang, Yaodong and Dong, Hao},
  booktitle=corl,
  year={2025}
}

@inproceedings{guzey2023dexterity,
  title={{Dexterity from Touch: Self-Supervised Pre-Training of Tactile Representations with Robotic Play}}, 
  author={Irmak Guzey and Ben Evans and Soumith Chintala and Lerrel Pinto},
  booktitle=corl,
  pages={3142--3166},
  year={2023},
  volume={229},
  series={Proceedings of Machine Learning Research},
  publisher={PMLR},
  url={https://proceedings.mlr.press/v229/guzey23a.html}
}

@inproceedings{DexPoint2022,
  title          = {{DexPoint: Generalizable Point Cloud Reinforcement Learning for Sim-to-Real Dexterous Manipulation}},
  author         = {Qin, Yuzhe and Huang, Binghao and Yin, Zhao-Heng and Su, Hao and Wang, Xiaolong},
  booktitle=corl,
  pages          = {594--605},
  year           = {2023},
  volume         = {205},
  series         = {Proceedings of Machine Learning Research},
  publisher      = {PMLR},
  url            = {https://proceedings.mlr.press/v205/qin23a.html}
}

@inproceedings{liu2025factr,
  title={{FACTR}: Force-Attending Curriculum Training for Contact-Rich Policy Learning},
  author={Jason Jingzhou Liu and Yulong Li and Kenneth Shaw and Tony Tao and Russ Salakhutdinov and Deepak Pathak},
  booktitle=rss,
  year={2025},
  address={Los Angeles, CA, USA},
  month={6},
  doi={10.15607/RSS.2025.XXI.079}
}

@inproceedings{Ze2024DP3,
  title={3D Diffusion Policy: Generalizable Visuomotor Policy Learning via Simple 3D Representations},
  author={Yanjie Ze and Gu Zhang and Kangning Zhang and Chenyuan Hu and Muhan Wang and Huazhe Xu},
  booktitle=rss,
  year={2024}
}

@inproceedings{chi2023diffusionpolicy,
  title={{Diffusion Policy: Visuomotor Policy Learning via Action Diffusion}},
  author={Chi, Cheng and Feng, Siyuan and Du, Yilun and Xu, Zhenjia and Cousineau, Eric and Burchfiel, Benjamin and Song, Shuran},
  booktitle=rss,
  year={2023}
}

@inproceedings{shaw2023leaphand,
  title={{LEAP Hand: Low-Cost, Efficient, and Anthropomorphic Hand for Robot Learning}},
  author={Shaw, Kenneth and Agarwal, Ananye and Pathak, Deepak},
  booktitle=rss,
  year={2023}
}

@inproceedings{Hao2024SopeDex,
  title = {{Learning to Singulate Objects in Packed Environments using a Dexterous Hand}},
  author = {Hao Jiang and Yuhai Wang and Hanyang Zhou and Daniel Seita}, 
  booktitle = isrr,
  year = {2024},
  url={https://sope-dex.github.io/}
}

@inproceedings{jiang2026dexmulti,
  title={{Concurrent Prehensile and Nonprehensile Manipulation: A Practical Approach to Multi-Stage Dexterous Tasks}},
  author={Hao Jiang and Yue Wu and Yue Wang and Gaurav S. Sukhatme and Daniel Seita},
  booktitle=iros,
  year={2026},
  url={https://dexmulti.github.io/}
}

@misc{lu2026modex,
  title={{MoDex: A Diffusion Policy for Sequential Multi-Object Dexterous Grasping}},
  author={Lu, Haofei and Liu, Hongjia and Dong, Yifei and Pokorny, Florian T. and Lundell, Jens and Kragic, Danica},
  year={2026},
  eprint={2606.05407},
  archivePrefix={arXiv},
  primaryClass={cs.RO}
}

@inproceedings{foong2026handful,
  title={{HANDFUL: Sequential Grasp-Conditioned Dexterous Manipulation with Resource Awareness}},
  author={Ethan Foong and Yunshuang Li and Hao Jiang and Gaurav S. Sukhatme and Daniel Seita},
  booktitle=iros,
  year={2026},
  url={https://handful-dex.github.io/}
}

@inproceedings{ferrari1992planning,
  title={{Planning Optimal Grasps}},
  author={Ferrari, Carlo and Canny, John},
  booktitle=icra,
  pages={2290--2295},
  year={1992},
  doi={10.1109/ROBOT.1992.219918}
}

@inproceedings{dogar2011framework,
  title={{A Framework for Push-Grasping in Clutter}},
  author={Dogar, Mehmet R. and Srinivasa, Siddhartha S.},
  booktitle=rss,
  year={2011},
  doi={10.15607/RSS.2011.VII.009}
}

@inproceedings{liu2023grounding,
  title={{Grounding DINO: Marrying DINO with Grounded Pre-Training for Open-Set Object Detection}},
  author={Liu, Shilong and Zeng, Zhaoyang and Ren, Tianhe and Li, Feng and Zhang, Hao and Yang, Jie and Li, Chunyuan and Yang, Jianwei and Su, Hang and Zhu, Jun and others},
  booktitle=eccv,
  year={2024}
}

@inproceedings{ravi2024sam,
  title={{SAM 2: Segment Anything in Images and Videos}},
  author={Ravi, Nikhila and Gabeur, Valentin and Hu, Yuan-Ting and Hu, Ronghang and Ryali, Chaitanya and Ma, Tengyu and Khedr, Haitham and R{\"a}dle, Roman and Rolland, Chloe and Gustafson, Laura and others},
  booktitle=iclr,
  year={2025}
}

@inproceedings{shi2026minimalist,
  title={Minimalist Compliance Control},
  author={Shi, Haochen and Hu, Songbo and Hou, Yifan and Wang, Weizhuo and Liu, C. Karen and Song, Shuran},
  booktitle=rss,
  year={2026}
}

@inproceedings{wu2024canonical,
  title={Canonical Representation and Force-Based Pretraining of {3D} Tactile for Dexterous Visuo-Tactile Policy Learning},
  author={Wu, Tianhao and Li, Jinzhou and Zhang, Jiyao and Wu, Mingdong and Dong, Hao},
  booktitle=icra,
  year={2025}
}

@inproceedings{li2025adaptacdex,
  title={Adaptive Visuo-Tactile Fusion with Predictive Force Attention for Dexterous Manipulation},
  author={Li, Jinzhou and Wu, Tianhao and Zhang, Jiyao and Chen, Zeyuan and Jin, Haotian and Wu, Mingdong and Shen, Yujun and Yang, Yaodong and Dong, Hao},
  booktitle=iros,
  year={2025}
}

@inproceedings{chen2025policyconsensus,
  title={Multi-Modal Manipulation via Multi-Modal Policy Consensus},
  author={Chen, Haonan and Xu, Jiaming and Chen, Hongyu and Hong, Kaiwen and Huang, Binghao and Liu, Chaoqi and Mao, Jiayuan and Li, Yunzhu and Du, Yilun and Driggs-Campbell, Katherine},
  booktitle=icra,
  year={2026}
}

@inproceedings{xue2025reactive,
  title={Reactive Diffusion Policy: Slow-Fast Visual-Tactile Policy Learning for Contact-Rich Manipulation},
  author={Xue, Han and Ren, Jieji and Chen, Wendi and Zhang, Gu and Fang, Yuan and Gu, Guoying and Xu, Huazhe and Lu, Cewu},
  booktitle=rss,
  year={2025}
}

@article{correll2018amazon,
  author={Correll, Nikolaus and Bekris, Kostas E. and Berenson, Dmitry and Brock, Oliver and Causo, Albert and Hauser, Kris and Okada, Kei and Rodriguez, Alberto and Romano, Joseph M. and Wurman, Peter R.},
  title={Analysis and Observations from the First Amazon Picking Challenge},
  journal={IEEE Transactions on Automation Science and Engineering},
  volume={15},
  number={1},
  pages={172--188},
  year={2018},
  doi={10.1109/TASE.2016.2600527}
}

@inproceedings{bai2026retrdex,
  author={Bai, Fengshuo and Li, Yu and Chu, Jie and Chou, Tawei and Zhu, Runchuan and Wen, Ying and Yang, Yaodong and Chen, Yuanpei},
  title={{RetrDex}: Efficient Object Retrieval in Cluttered Scenes with a Dexterous Hand},
  booktitle=iros,
  year={2026}
}

@inproceedings{zhang2026cadgrasp,
  author={Zhang, Jiyao and Ma, Zhiyuan and Wu, Tianhao and Chen, Zeyuan and Dong, Hao},
  title={{CADGrasp}: Learning Contact and Collision Aware General Dexterous Grasping in Cluttered Scenes},
  booktitle=neurips,
  year={2025}
}

@inproceedings{mahler2017dexnet,
  author={Jeffrey Mahler and Jacky Liang and Sherdil Niyaz and Michael Laskey and Richard Doan and Xinyu Liu and Juan Aparicio and Ken Goldberg},
  title={{Dex-Net 2.0}: Deep Learning to Plan Robust Grasps with Synthetic Point Clouds and Analytic Grasp Metrics},
  booktitle=rss,
  year={2017},
  address={Cambridge, MA, USA},
  month={7},
  doi={10.15607/RSS.2017.XIII.058}
}

@inproceedings{fang2020graspnet,
  author={Fang, Hao-Shu and Wang, Chenxi and Gou, Minghao and Lu, Cewu},
  title={{GraspNet-1Billion}: A Large-Scale Benchmark for General Object Grasping},
  booktitle=cvpr,
  pages={11444--11453},
  month={6},
  year={2020},
  url={https://openaccess.thecvf.com/content_CVPR_2020/html/Fang_GraspNet-1Billion_A_Large-Scale_Benchmark_for_General_Object_Grasping_CVPR_2020_paper.html}
}

@article{kappassov2015tactile,
  title={Tactile Sensing in Dexterous Robot Hands: Review},
  volume={74},
  number={Part A},
  journal={Robotics and Autonomous Systems},
  author={Kappassov, Zhanat and Corrales, Juan-Antonio and Perdereau, V{\'e}ronique},
  year={2015},
  month={12},
  pages={195--220},
  doi={10.1016/j.robot.2015.07.015}
}

@article{calandra2018more,
  title={More Than a Feeling: Learning to Grasp and Regrasp Using Vision and Touch},
  volume={3},
  number={4},
  journal={IEEE Robotics and Automation Letters},
  author={Calandra, Roberto and Owens, Andrew and Jayaraman, Dinesh and Lin, Justin and Yuan, Wenzhen and Malik, Jitendra and Adelson, Edward H. and Levine, Sergey},
  year={2018},
  month={10},
  pages={3300--3307},
  doi={10.1109/LRA.2018.2852779}
}

@inproceedings{qi2023rotateit,
  title={General In-Hand Object Rotation with Vision and Touch},
  author={Qi, Haozhi and Yi, Brent and Suresh, Sudharshan and Lambeta, Mike and Ma, Yi and Calandra, Roberto and Malik, Jitendra},
  booktitle=corl,
  pages={2549--2564},
  year={2023},
  volume={229},
  series={Proceedings of Machine Learning Research},
  publisher={PMLR},
  url={https://proceedings.mlr.press/v229/qi23a.html}
}

@inproceedings{chen2023vtt,
  title={Visuo-Tactile Transformers for Manipulation},
  author={Chen, Yizhou and Van der Merwe, Mark and Sipos, Andrea and Fazeli, Nima},
  booktitle=corl,
  pages={2026--2040},
  year={2023},
  volume={205},
  series={Proceedings of Machine Learning Research},
  publisher={PMLR},
  url={https://proceedings.mlr.press/v205/chen23d.html}
}

@inproceedings{huang2025vitac,
  title={{3D-ViTac}: Learning Fine-Grained Manipulation with Visuo-Tactile Sensing},
  author={Huang, Binghao and Wang, Yixuan and Yang, Xinyi and Luo, Yiyue and Li, Yunzhu},
  booktitle=corl,
  pages={2557--2578},
  year={2025},
  volume={270},
  series={Proceedings of Machine Learning Research},
  publisher={PMLR},
  url={https://proceedings.mlr.press/v270/huang25e.html}
}

@inproceedings{heng2025vitacformer,
  title={{ViTacFormer}: Learning Cross-Modal Representation for Visuo-Tactile Dexterous Manipulation},
  author={Heng, Liang and Geng, Haoran and Zhang, Kaifeng and Abbeel, Pieter and Malik, Jitendra},
  year={2026},
  booktitle=rss,
}

@inproceedings{higuera2025sparshx,
  title={Tactile Beyond Pixels: Multisensory Touch Representations for Robot Manipulation},
  author={Higuera, Carolina and Sharma, Akash and Fan, Taosha and Bodduluri, Chaithanya Krishna and Boots, Byron and Kaess, Michael and Lambeta, Mike and Wu, Tingfan and Liu, Zixi and Hogan, Francois Robert and Mukadam, Mustafa},
  booktitle=corl,
  year={2025}
}

@misc{huang2025sata,
  title={Spatially anchored Tactile Awareness for Robust Dexterous Manipulation},
  author={Huang, Jialei and Ye, Yang and Gong, Yuanqing and Zhu, Xuezhou and Gao, Yang and Zhang, Kaifeng},
  year={2025},
  eprint={2510.14647},
  archivePrefix={arXiv},
  primaryClass={cs.RO},
  url={https://arxiv.org/abs/2510.14647}
}

@misc{niu2026trex,
  title={{T-Rex: Tactile-Reactive Dexterous Manipulation}},
  author={Niu, Dantong and Liu, Zhuoyang and Wang, Zekai and Shao, Boning and Yin, Zhao-Heng and Pai, Anirudh and Sharma, Yuvan and Saravalle, Stefano and Zheng, Ruijie and Wang, Jing and Punamiya, Ryan and Xu, Mengda and Xie, Yuqi and Jiang, Yunfan and Fu, Letian and Kallidromitis, Konstantinos and Gioia, Matteo and Zhang, Junyi and Ge, Jiaxin and Feng, Haiwen and Galasso, Fabio and Zhan, Wei and Chan, David M. and Bai, Yutong and Herzig, Roei and Lei, Jiahui and Fei-Fei, Li and Goldberg, Ken and Malik, Jitendra and Abbeel, Pieter and Zhu, Yuke and Xu, Danfei and Fan, Linxi and Darrell, Trevor},
  year={2026},
  eprint={2606.17055},
  archivePrefix={arXiv},
  primaryClass={cs.RO}
}

@misc{ke2026tacdexgrasp,
  title={{TacDexGrasp: Compliant and Robust Dexterous Grasping with Tactile Feedback}},
  author={Ke, Yubin and Chen, Jiayi and Lv, Hang and Zhou, Xiao and Wang, He},
  year={2026},
  eprint={2603.07040},
  archivePrefix={arXiv},
  primaryClass={cs.RO}
}

@inproceedings{xu2026cgp,
  title={{Contact-Grounded Policy: Dexterous Visuotactile Policy with Generative Contact Grounding}},
  author={Xu, Zhengtong and Wang, Yeping and Abbatematteo, Ben and Preechayasomboon, Jom and Chan, Sonny and Colonnese, Nick and Memar, Amirhossein H.},
  year={2026},
  booktitle=rss,
}

@inproceedings{higuera2024sparsh,
  title={{Sparsh: Self-supervised Touch Representations for Vision-based Tactile Sensing}},
  author={Higuera, Carolina and Sharma, Akash and Bodduluri, Chaithanya Krishna and Fan, Taosha and Lancaster, Patrick and Kalakrishnan, Mrinal and Kaess, Michael and Boots, Byron and Lambeta, Mike and Wu, Tingfan and Mukadam, Mustafa},
  booktitle=corl,
  year={2024}
}

@inproceedings{zhang2026touchguide,
  title={{TouchGuide: Inference-Time Steering of Visuomotor Policies via Touch Guidance}},
  author={Zhang, Zhemeng and Ma, Jiahua and Yang, Xincheng and Wen, Xin and Zhang, Yuzhi and Li, Boyan and Qin, Yiran and Liu, Jin and Zhao, Can and Kang, Li and Hong, Haoqin and Yin, Zhenfei and Torr, Philip and Su, Hao and Zhang, Ruimao and Ma, Daolin},
  booktitle=rss,
  year={2026}
}

@misc{gupta2026lucid,
  title={{LUCID: Learning Embodiment-Agnostic Intent Models from Unstructured Human Videos for Scalable Dexterous Robot Skill Acquisition}},
  author={Gupta, Harsh and Shi, Guanya and Yuan, Wenzhen},
  year={2026},
  eprint={2606.11628},
  archivePrefix={arXiv},
  primaryClass={cs.RO}
}

@misc{ma2026current,
  title={{Current as Touch: Proprioceptive Contact Feedback for Compliant Dexterous Manipulation}},
  author={Ma, Chenyang and Yao, Yunchao and Wei, Zhenyu and Li, Ruogu and Szafir, Daniel and Ding, Mingyu},
  year={2026},
  eprint={2607.03529},
  archivePrefix={arXiv},
  primaryClass={cs.RO}
}

@inproceedings{dou2026neuralactuator,
  title={{NeuralActuator: Neural Actuation Modeling for Robot Dynamics and External Force Perception}},
  author={Dou, Zhiyang and Onyemelukwe, John U. and Zhang, Hangxing and Zhang, Heng and Guo, Minghao and Tian, Yunsheng and Lipiec, Michal Piotr and Jacob, Joshua and Liu, Chao and Chen, Peter Yichen and Ivanov, Yuri and Matusik, Wojciech},
  booktitle=rss,
  year={2026}
}

@misc{wu2026reforce,
  title={{ReForce: Learning Force-aware Retargeting for Dexterous Manipulation}},
  author={Wu, Yuhang and Zeng, Lingqi and Jing, Changwei and Ye, Jianglong and Wang, Xiaolong},
  year={2026},
  eprint={2608.15560},
  archivePrefix={arXiv},
  primaryClass={cs.RO}
}

@misc{zhu2026chord,
  title={{Learning Dexterous Manipulation Using Contact Wrench Guidance From Human Demonstration}},
  author={Zhu, Xinghao and Liu, Zixi and Jain, Shalin and Li, Chenran and Noori, Milad and Lin, Michael Andres and Zhao, Huihua and Welsh, John and Verghese, Mrinal and Liu, Wei and Wang, Tingwu and Da, Xingye and Luo, Zhengyi and Kulkarni, Vishal and Bhatti, Naema and Zhu, Yuke and Fan, Linxi and Wen, Bowen and Xu, Danfei and Pouya, Soha and Chang, Yan},
  year={2026},
  eprint={2607.00033},
  archivePrefix={arXiv},
  primaryClass={cs.RO}
}

@misc{lum2026play2perfect,
  title={{Play2Perfect: What Matters in Dexterous Play Pretraining for Precise Assembly?}},
  author={Lum, Tyler Ga Wei and Kedia, Kushal and Liu, C. Karen and Bohg, Jeannette},
  year={2026},
  eprint={2606.26428},
  archivePrefix={arXiv},
  primaryClass={cs.RO}
}

@inproceedings{zheng2026dapl,
  title={{Emerging Extrinsic Dexterity in Cluttered Scenes via Dynamics-aware Policy Learning}},
  author={Zheng, Yixin and Lyu, Jiangran and Zhang, Yifan and Chen, Jiayi and Yan, Mi and Deng, Yuntian and Shi, Xuesong and Zhao, Xiaoguang and Wang, Yizhou and Zhang, Zhizheng and Wang, He},
  year={2026},
  booktitle=rss
}
